\documentclass[10pt,twocolumn,letterpaper]{article}

\usepackage{cvpr}              % To produce the CAMERA-READY version
\usepackage[accsupp]{axessibility}
\graphicspath{{pdf/}}
\usepackage{makecell}
\usepackage{multirow}
\usepackage{color}
\definecolor{Y}{RGB}{230,220,180}
\definecolor{B}{RGB}{172,189,215}
\usepackage{mathrsfs}
\usepackage{xcolor}
\usepackage{amsmath, amssymb}
\usepackage{booktabs}
\usepackage{cuted}   % 允许临时单栏/双栏切换
\usepackage{caption} % 支持 captionof
\definecolor{deepskyblue}{RGB}{0,191,255} 
\definecolor{darkyellow}{RGB}{204,153,0} % 自定义深黄色
\definecolor{lightred}{RGB}{255,102,102} % 自定义浅红色
\usepackage{marvosym}

\definecolor{cvprblue}{rgb}{0.21,0.49,0.74}
\usepackage[pagebackref,breaklinks,colorlinks,allcolors=cvprblue]{hyperref}

\def\paperID{41245} % *** Enter the Paper ID here
\def\confName{CVPR}
\def\confYear{2026}

\title{FontCrafter: High-Fidelity Element-Driven Artistic Font Creation with Visual In-Context Generation}

\author{
Wuyang Luo$^1$\textsuperscript{\Letter} \qquad
Chengkai Tan$^1$ \qquad
Chang Ge$^1$ \qquad
Binye Hong$^1$ \qquad
Su Yang$^2$ \qquad
Yongjiu Ma$^1$\\
{\small$^1$Dalian University of Technology} \qquad 
{\small$^2$Shanghai Key Laboratory of Intelligent Information Processing, Fudan University}\\
}

\begin{document}
\maketitle

\let\thefootnote\relax
\footnotetext{\textsuperscript{\Letter} Corresponding author: Wuyang Luo}

% % ---------- Teaser Figure (full-width, fixed position) ----------
\begin{strip}
    \centering
    \vspace{-5em}
    \includegraphics[width=\textwidth]{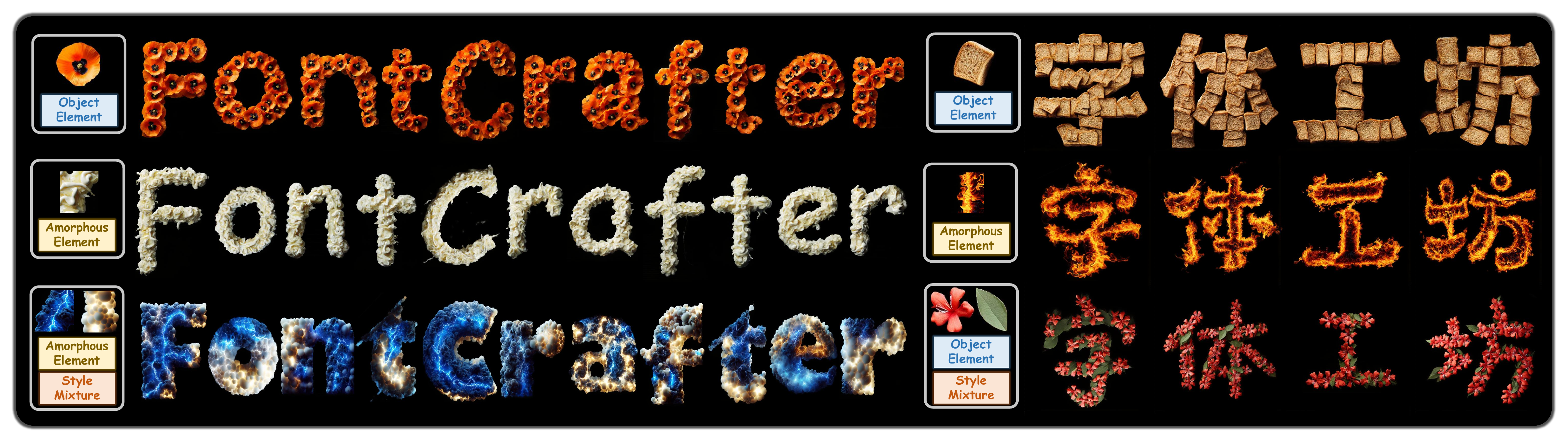}
    \vspace{-2em}
    \captionof{figure}{Our artistic font generation system produces high-fidelity stylized glyphs driven by reference elements, supporting both \textbf{\textcolor{deepskyblue}{Object Elements}} and \textbf{\textcolor{darkyellow}{Amorphous Elements}}. The generated glyphs preserve the textures and structural characteristics of the reference elements while maintaining strong style consistency across glyphs. Our method also enables flexible \textbf{\textcolor{lightred}{Style Mixture}} from multiple elements.}
    \label{fig:teaser}
    % \vspace{-0.9em}
\end{strip}
% % ---------------------------------------------------------------

\begin{abstract}
Artistic font generation aims to synthesize stylized glyphs based on a reference style. However, existing approaches suffer from limited style diversity and coarse control. In this work, we explore the potential of element-driven artistic font generation. Elements are the fundamental visual units of a font, serving as reference images for the desired style. Conceptually, we categorize elements into object elements (e.g., flowers or stones) with distinct structures and amorphous elements (e.g., flames or clouds) with unstructured textures. We introduce FontCrafter, an element-driven framework for font creation, and construct a large-scale dataset, ElementFont, which contains diverse element types and high-quality glyph images. However, achieving high-fidelity reconstruction of both texture and structure of reference elements remains challenging. To address this, we propose an in-context generation strategy that treats element images as visual context and uses an inpainting model to transfer element styles into glyph regions at the pixel level. To further control glyph shapes, we design a lightweight Context-aware Mask Adapter (CMA) that injects shape information. Moreover, a training-free attention redirection mechanism enables region-aware style control and suppresses stroke hallucination. In addition, edge repainting is applied to make boundaries more natural. Extensive experiments demonstrate that FontCrafter achieves strong zero-shot generation performance, particularly in preserving structural and textural fidelity, while also supporting flexible controls such as style mixture. 

\end{abstract}    
\section{Introduction}
\label{sec:intro}

\begin{figure}[t]
    \centering
    \includegraphics[width=8.3cm, trim=10 10 10 10,clip]{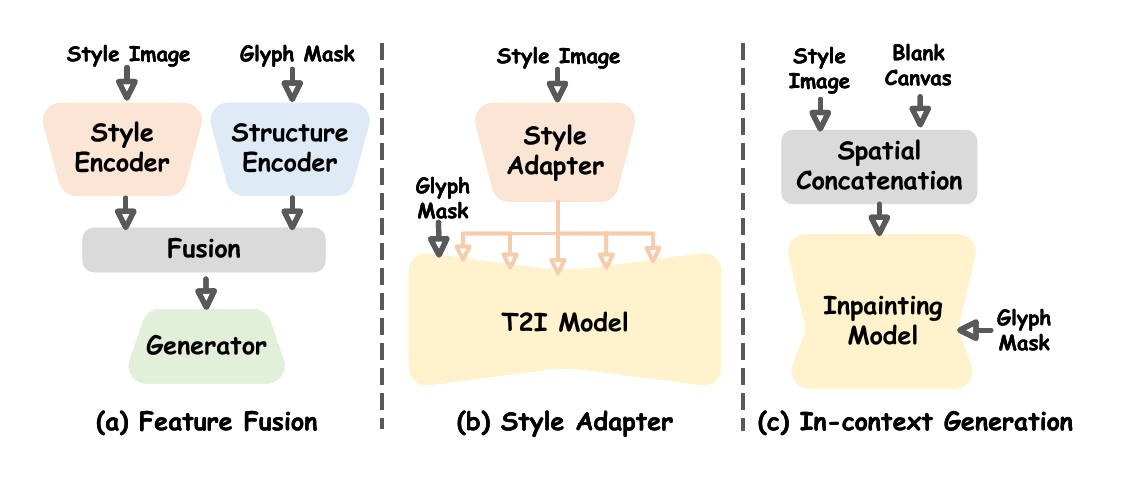}
    \caption{Comparison of different font style control strategies.}
    \label{fig:intro}
\end{figure}

Artistic font generation aims to synthesize stylized glyphs from a user-defined reference image and glyph mask, requiring precise style transfer and structural alignment. Existing methods fall into two main paradigms. As shown in Figure \ref{fig:intro}(a), the first follows a feature fusion strategy \cite{yang2019tet,li2020fet,li2023compositional}, typically built on Generative Adversarial Networks (GANs) \cite{goodfellow2020generative}. These methods utilize separate encoders to extract representations from the style and glyph images, fuse them in the feature space, and generate stylized glyphs. However, limited model capacity and small-scale training data with simple textures often lead to unrealistic results, poor adaptation to complex styles, and weak generalization to unseen references. The second paradigm \cite{mu2024fontstudio,shi2025fonts}, illustrated in Figure \ref{fig:intro}(b), leverages pretrained text-to-image diffusion models with adapter modules to enable zero-shot generation. Style conditions are injected through a style adapter, such as IP-Adapter \cite{ye2023ip}. However, these methods often fail to produce results that closely match the reference style, as style adapters capture only global features and overlook pixel-level details. In addition, most existing methods support only coarse-grained controls, such as color or overall style, and cannot meet diverse user requirements.

To address the limited style diversity and controllability of existing methods, we reformulate reference styles using the concept of elements. Elements are categorized into two types: amorphous elements, characterized by texture-like patterns, and object elements, consisting of distinct instance-level objects. To support element-driven control, we construct ElementFont, a large-scale dataset specifically designed for artistic font generation. Each font image is paired with its corresponding element, and these fine-grained annotations enhance controllability while enabling broader applications.

\begin{figure}[t]
    \centering
    \includegraphics[width=6.3cm, trim=10 10 10 10,clip]{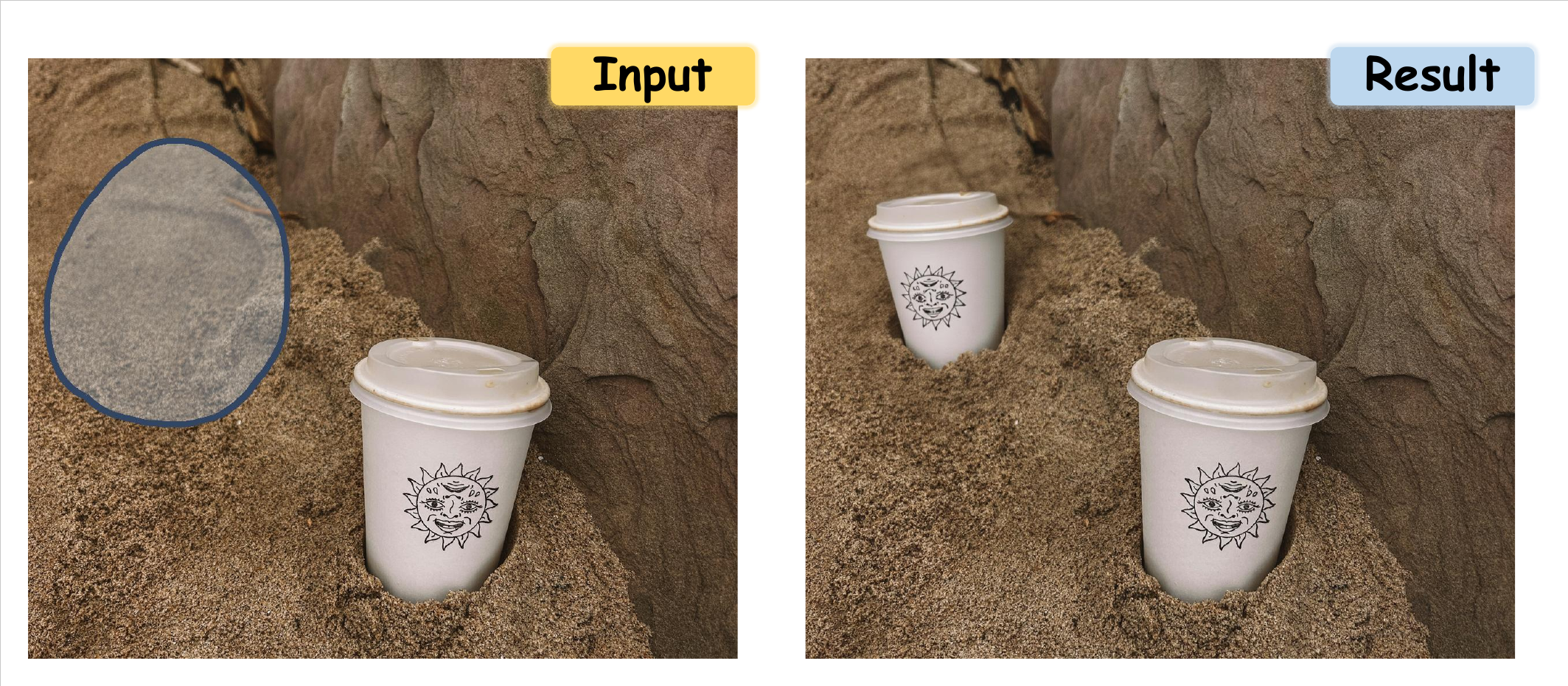}
    \caption{Context transfer in image inpainting (e.g., FLUX.1-Fill). 
When an input image contains a white cup with a masked region, prompting the model with ``a cup'' enables it to reconstruct the masked region by propagating visual cues from the visible context. This property motivates our formulation of artistic font generation as visual in-context generation.}
    \label{fig:sm_motivation}
\end{figure}

To achieve high-fidelity generation that preserves the visual appearance of elements, we draw inspiration from the context-transfer capability of recent image inpainting models, such as FLUX.1-Fill \cite{fluxfill}, which utilizes contextual pixels to complete missing regions, as illustrated in Figure~\ref{fig:sm_motivation}. 
Motivated by this insight, we propose FontCrafter, which formulates our task as visual in-context generation. Specifically, element images are concatenated with a blank canvas in pixel space and fed into a pretrained inpainting model. The element region serves as visual context, while the glyph region is treated as the masked region, allowing the model to transfer element styles onto glyphs. Finally, we introduce three components to enhance generation quality and controllability. The Context-aware Mask Adapter (CMA) fuses the glyph mask with contextual features to produce adaptive shape-control signals that inject glyph structure. An attention redirection module manipulates self-attention to suppress stroke hallucination and enable region-aware style mixture. In addition, an edge repainting stage refines glyph boundaries for more natural alignment with reference elements. Together, these components enable FontCrafter to generate high-fidelity element-driven fonts while supporting controllable generation, such as style mixture. Our main contributions are summarized as follows: 
\begin{itemize}
\item We propose FontCrafter, an element-driven artistic font generation framework via visual in-context generation, enabling high-fidelity style transfer with flexible control.
\item We construct ElementFont, a large-scale dataset for element-driven artistic font generation with diverse element types.
\item Extensive experiments demonstrate that FontCrafter achieves high-fidelity font generation while faithfully preserving the texture and structure of reference elements.
\end{itemize}

\section{Related Works}

\textbf{Artistic Font Generation.}
Artistic font generation aims to synthesize stylized characters by transferring visual patterns from a reference image onto a given glyph, while preserving legibility. Transferable styles span two dimensions: glyph topology and visual texture. The first line of work focuses on structural deformation to emulate diverse stroke styles. Early approaches \cite{tian2017zi2zi,chang2018chinese,jiang2017dcfont,sun2018pyramid,lyu2017auto}. formulate this task as image-to-image translation, learning mappings between different font domains from paired datasets. To improve generalization and reduce data dependence, later works explore few-shot strategies, including content-style disentanglement \cite{sun2018learning,zhang2018separating,yang2024fontdiffuser,wang2023cf,gao2019artistic}, stroke-level supervision \cite{gao2020gan,liu2022xmp,he2024diff}, glyph structure annotations \cite{jiang2019scfont,park2021few}, and unsupervised learning \cite{xie2021dg}. Some methods further extend stroke deformation to semantic typography \cite{iluz2023word,he2023wordart}.   
The second line focuses on transferring texture patterns from reference images onto glyph masks \cite{yang2017awesome,azadi2018multi,yang2018context,yang2019tet,yang2019controllable,li2020fet,li2023compositional}. Early methods \cite{yang2017awesome,yang2018context} use patch-based matching, while GAN-based approaches like MC-GAN \cite{azadi2018multi}, TET-GAN \cite{yang2019tet}, and Shape-Matching GAN \cite{yang2019controllable} enhance few-shot style transfer and controllability. Recently, diffusion models have emerged: Anything2Glyph \cite{wang2023anything} leverages a text-to-image diffusion model to generate glyphs composed of objects under prompt guidance. Similarly, FontStudio \cite{mu2024fontstudio} introduces a shape-adaptive diffusion model for producing coherent and consistent stylized fonts. FonTS \cite{shi2025fonts} renders stylized text guided by reference topology and texture.  
Our work follows this line of research. However, unlike diffusion methods relying solely on textual prompts, we condition on reference elements, enabling fine-grained control over visual appearance and supporting both amorphous and object elements, thereby enhancing the diversity and expressiveness of controllable styles.

% Our work follows this category, but unlike diffusion methods relying solely on textual prompts, we condition on reference elements, enabling fine-grained control over visual appearance and supporting both amorphous and object elements, thus enhancing diversity and expressiveness of controllable signals.

\noindent
\textbf{Controllable Generation in Diffusion Models.}
Text-to-image diffusion models \cite{rombach2022high,podell2023sdxl} establish language–vision alignment, surpassing GAN-based methods in generation \cite{luo2023reference, luo2022photo, xu2025b4m, xu2025context,wang2024oneactor, wang2025spotactor,zhou2026spatialrewardverifiablespatialreward} and editing \cite{luo2022context, luo2023siedob, xu2024headrouter, xu2026tag, zhou2026unified}, and can serve as a data engine to support tasks across various other domains \cite{song2025hume, xu2026beyond, wang2025spatialclip}. 
Recently, various methods introduce additional control signals for task-specific image generation. Paint by Example \cite{yang2023paint} enables exemplar-guided editing. ControlNet \cite{zhang2023adding} provides pixel-level spatial alignment via conditional branches. T2I-Adapter \cite{mou2024t2i} improves conditioning efficiency with spatial adapters. UniControl \cite{qin2023unicontrol} and Uni-ControlNet \cite{zhao2023uni} unify diverse spatial conditions. IP-Adapter \cite{ye2023ip} injects style features from reference images via cross-attention, eliminating the need for explicit spatial alignment. MimicBrush \cite{chen2024zero} transfers cues from exemplar images for reference-guided generation. Recent image editing methods \cite{chong2024catvton,zhangenabling,luo2026soedit} employ the original image as a conditioning signal to control the editing results. 
Despite success in natural images, these methods struggle with artistic font synthesis due to the substantial domain gap. Here, we address style controllability with an element-driven framework that leverages diverse visual references to generate glyphs with flexible, fine-grained control.

\section{ElementFont Dataset}

\begin{figure}[t]
    \centering
    \includegraphics[width=8.0cm, trim=10 10 10 10,clip]{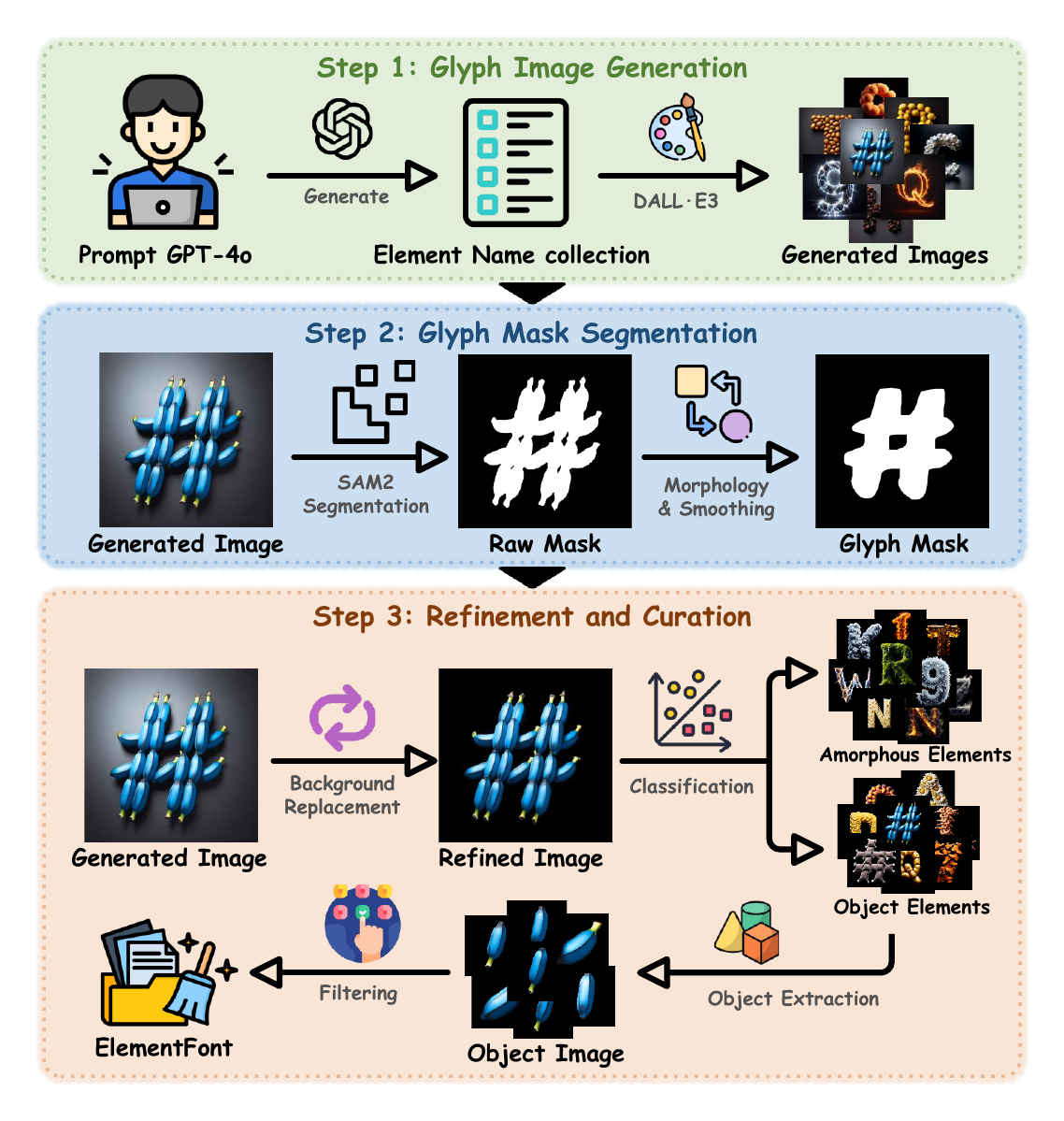}
    \caption{Dataset collection and construction pipeline.}
    \label{fig:dataset_pipeline}
\end{figure}

\begin{figure*}[t]
    \centering
    \includegraphics[width=16.5cm, trim=10 10 10 10,clip]{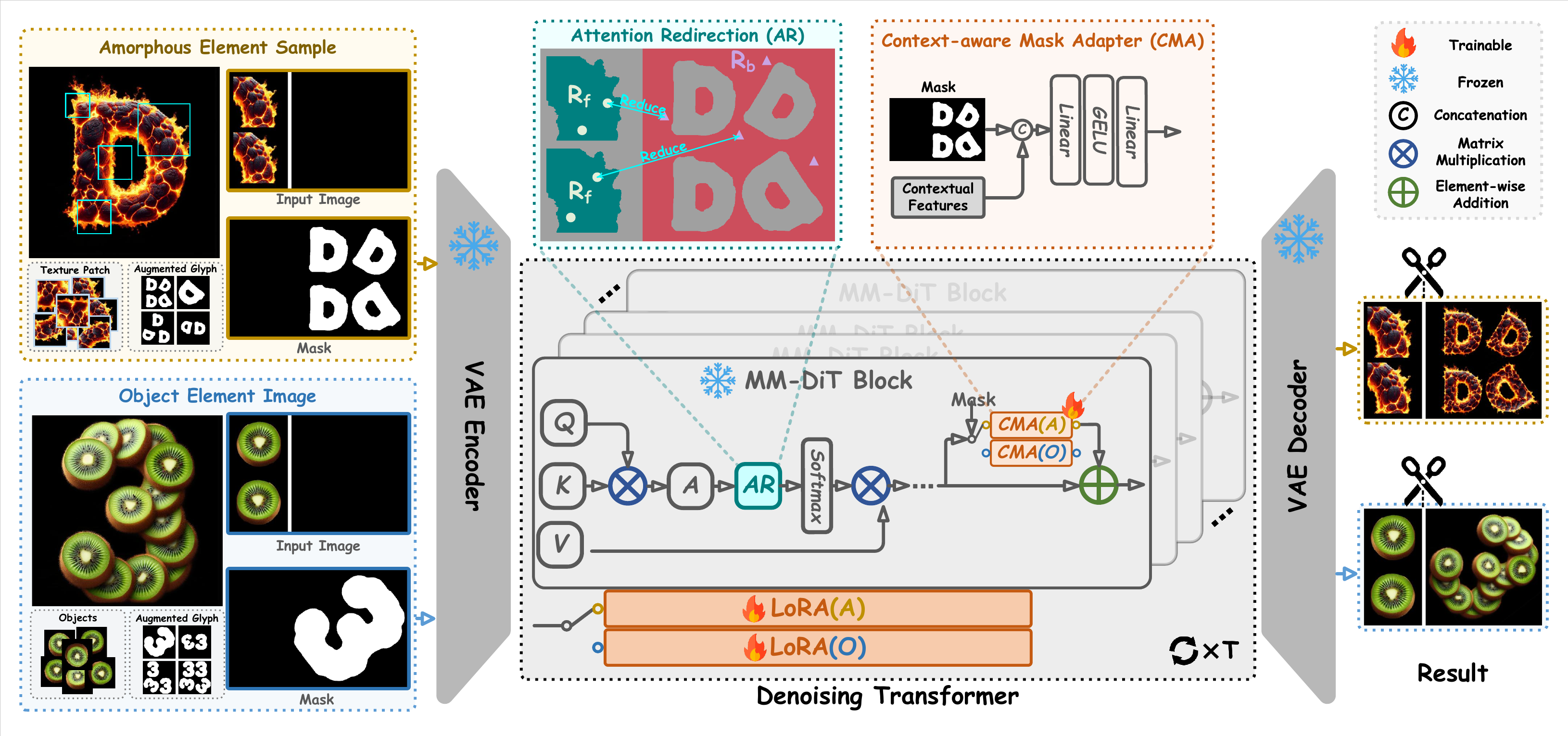}
    \caption{Overview of the proposed FontCrafter framework. We spatially combine the elements with a blank canvas as input, then apply the Context-aware Mask Adapter (CMA) for glyph control and use Attention Redirection (AR) to regulate the spatial influence of the elements.}
    \label{fig:framework}
\end{figure*}

The lack of large-scale artistic font datasets remains a major bottleneck. Existing methods often overlook the diversity of reference styles and rely on simple texture patches as style sources. We observe that for basic glyph categories such as English letters, digits, and punctuation, commercial tools like DALL·E 3 can generate high-quality stylized images, enabling the collection of foundational training samples. In addition, we categorize reference images into amorphous and object types and construct ElementFont, a large-scale dataset featuring diverse elements. The automated data generation pipeline is illustrated in Figure \ref{fig:dataset_pipeline}.

\noindent
\textbf{Glyph Image Generation.}
We begin by using a large language model to generate a diverse collection of element names, which specify the visual styles of glyphs. These elements are crafted to be both visually distinctive and suitable for stylized font generation. To promote stylistic diversity and fine-grained categorization, we prompt GPT-4o to produce element names in the form of \textit{"adjective + noun"}, such as \textit{"blue java banana"}. Each name is then inserted into a fixed textual template: \textit{"Character [C] made up of [E] with a pure black background"}, where \textit{[C]} denotes the target character and \textit{[E]} refers to the selected element. Finally, we use DALL·E 3 to generate multiple stylized glyph images for each element, resulting in a total of 32,000 raw samples.

\noindent
\textbf{Glyph Mask Segmentation.}
To obtain paired glyph images and masks, we use SAM2 \cite{ravi2024sam} to segment glyph regions and generate initial binary masks. However, these masks are tightly aligned with the contours of stylized glyphs and often incorporate visual features specific to the reference elements. In contrast, glyph masks used during inference typically come from standard font libraries and feature clean contours without stylistic distortions. To bridge this gap, we refine the initial masks using a combination of morphological operations and contour-based Gaussian smoothing, producing clean masks consistent with real-world usage.

\noindent
\textbf{Refinement and Curation.}
Although paired data is available at this stage, further processing is required to ensure high-quality training samples. (1) Background Replacement: Although DALL·E 3 is instructed to generate images with pure black backgrounds, many samples contain undesired lighting artifacts and shadows. We extract glyph foregrounds using the raw masks and place them onto clean black backgrounds. However, this process often introduces unnatural edges. To address this issue, we train a dedicated inpainting model to restore boundary regions with visual coherence. (2) Classification and Extraction: We divide glyph samples into two categories: amorphous elements and object elements. For samples in the object category, we further extract instance-level masks corresponding to the embedded objects. (3) Sample Filtering: To ensure dataset correctness, we use GPT to inspect each sample and remove those containing errors in any component, including the glyph image, glyph mask, or object mask. Samples with inaccurate or misaligned components are discarded. In total, the dataset contains 14000 glyphs in the Amorphous category and 5000 in the Object category, covering 6000 distinct element types. 

\section{FontCrafter}
% \label{sec:formatting}

The goal of this work is to develop a zero-shot artistic font generation model, FontCrafter, which synthesizes stylized character images conditioned on two inputs: a reference element image and a glyph mask, both unseen during training. The generated result is expected to faithfully reflect the visual style conveyed by the element while remaining topologically consistent with the glyph mask.

The diffusion-based image inpainting model, FLUX.1-Fill \cite{fluxfill}, achieves strong performance in completing masked regions with high visual fidelity by leveraging surrounding context. This capability is largely enabled by stacked MultiModal DiT (MM-DiT) blocks \cite{peebles2023scalable}, which aggregate contextual information across tokens via attention mechanism, allowing visual content to propagate from visible to masked regions. 
Each MM-DiT block performs self-attention over a joint sequence of text and image tokens. The query, key, and value matrices $Q, K, V \in \mathbb{R}^{L \times d_k}$ are formed by concatenating text and image token embeddings. The attention output is computed as $\operatorname{Attention}(Q, K, V)=A V=\operatorname{softmax}(Q K^T/\sqrt{d_k}) V$, where $A \in \mathbb{R}^{L \times L}$ denotes the attention weight matrix. Each row of $A$ represents a normalized distribution over all tokens.

Inspired by the strong visual content transfer capability of FLUX.1-Fill, which propagates appearance cues from visible to masked regions, we reformulate artistic font generation as a visual in-context generation task to naturally transfer element styles onto target glyphs. The overall pipeline of the proposed method is illustrated in Figure~\ref{fig:framework}. Our framework is built upon FLUX.1-Fill, with task-irrelevant architectural details omitted. 
Specifically, we spatially concatenate the element image with a blank canvas, which is an all-zero image of the same size as the glyph mask, to construct the input. To explicitly indicate the glyph shape and guide the generation process, we insert a Context-aware Mask Adapter (CMA) into each MM-DiT block.

\noindent
\textbf{Context-aware Mask Adapter.}
Our proposed Context-aware Mask Adapter (CMA) is a simple and lightweight module designed to generate shape-aware control signals, as illustrated in the top-right of Figure~\ref{fig:framework}. It is inserted at the end of each MM-DiT block and consists of two linear layers with a GELU activation in between. The first linear layer reduces the channel dimension to 64, and the second restores it to the original dimension. The CMA input is formed by concatenating the downsampled glyph mask with the output features of the MM-DiT block along the channel dimension. These features contain contextual information from the generation process and structural cues derived from the reference element. If the glyph mask alone were used to generate control signals, the resulting features would be independent of the reference element. However, even for the same glyph, different reference elements should result in distinct element-aware structural characteristics. By fusing contextual features with the glyph mask, CMA can adaptively generate control signals conditioned on different inputs, enabling the model to capture element-specific structures consistent with the reference element.

\noindent
\textbf{Model Training.}
To construct triplet training data comprising the input image $I_{input}$, the glyph mask $I_{glyph}$, and the ground truth $I_{gt}$, we first prepare two components: a reference region and a glyph region. Given a glyph image with resolution $H \times W$, we create a reference region of size $H \times \frac{W}{2}$. For amorphous elements, we randomly crop a texture patch centered within the glyph area and vertically stack two such patches to fill the reference region as fully as possible. For object elements, we randomly select several segmented object instances from the glyph image and concatenate them to form the reference region. The glyph region is constructed by combining one to four original glyph masks, each randomly rotated. This augmentation compensates for the limited diversity of the training set, which primarily contains simple Latin glyphs, whereas real-world applications, such as Chinese character synthesis, involve far more complex structures. By introducing glyph composition and rotation, we increase structural diversity and complexity, thereby enhancing the model’s zero-shot generalization. 
After generating both the reference and glyph regions, we horizontally concatenate them to form the ground truth $I_{gt}$. Similarly, the model input $I_{input}$ is constructed by concatenating the reference region with a blank canvas, while the glyph mask $I_{glyph}$ is obtained by concatenating an all-zero region of the same size as the reference region with the glyph regions. Direct concatenation of the two parts may cause visual artifacts, such as blending or unintended connections between regions. To prevent this, we insert a narrow separation band of width $H \times \frac{W}{32}$ between the reference and glyph regions, ensuring that the generated output remains cleanly separated and that the stylized glyph can be easily extracted by cropping the corresponding area. 

We fine-tune the denoising transformer using LoRA \cite{hu2022lora} on all linear layers within each MM-DiT block. The CMA modules are trained jointly with LoRA, while all other model parameters remain frozen to ensure parameter-efficient adaptation. Due to the substantial differences between amorphous and object elements, we use independent LoRA and CMA parameters for each element type. Since the trainable parameters account for only 0.5\% of the entire model, switching adapters between element types incurs negligible computational overhead. The model is optimized using the flow matching loss \cite{esser2024scaling} with a learning rate of $1\times10^{-4}$. Because the reference image provides sufficient style conditioning, the text input is set to empty during training. During inference, the model can generate stylized glyphs conditioned on unseen elements and arbitrary glyphs.

\noindent
\textbf{Attention Redirection.}
We introduce Attention Redirection for two main purposes. First, the model occasionally generates extraneous content outside the glyph regions, causing structural errors. Attention Redirection mitigates such hallucinations by suppressing unintended structures in the background. Second, it enables region-aware control during multi-element style mixing. Specifically, we define a suppression matrix $M_{attenuate} \in \mathbb{R}^{L \times L}$ matching the dimensions of the attention map $A$: 
\begin{equation}
M_{\text {attenuate }}(i, j)=\left\{\begin{array}{cc}
1, & \text { if token } \mathrm{i} \in R_b \text { and } \mathrm{j} \in R_f \\
0, & \text { otherwise }
\end{array}\right.
\end{equation}
Here, $R_b$ and $R_f$ denote the background region of the glyph and the foreground region of the reference, respectively. During self-attention computation, we modify the attention logits to suppress cross-region interactions as follows: 
\begin{equation}
\hat{A}=A+M_{\text {attenuate }} \cdot \log _e(\lambda)
\end{equation}
\begin{equation}
\text { Attention }(Q, K, V)=\operatorname{softmax}(\hat{A}) V
\end{equation}
Where $\lambda \in(0,1)$ is a suppression factor. Since $\log _e(\lambda)<0$, the attention from reference foreground tokens to glyph background tokens is downweighted. For a token pair $(i,j)$ with $M_{\text {attenuate }}(i, j)=1$, the adjusted attention becomes: 
\begin{equation}
\operatorname{softmax}(\hat{A}_{i,j}) = \frac{e^{A_{i,j} + \log_e(\lambda)}}{\sum_k e^{\hat{A}_{i,k}}} = \frac{\lambda \cdot e^{A_{i,j}}}{\sum_k e^{\hat{A}_{i,k}}}
\end{equation}
This reduces the original attention weight by a factor of $\lambda$, preventing undesired strokes in the background. Suppression is applied exclusively to image–image token pairs, leaving text-related attentions unchanged. By integrating this mechanism into each attention layer, foreground pixels in the reference image are prevented from influencing the glyph background, encouraging the model to transfer style only to the masked stroke regions and improving structural consistency of the generated glyph.

\begin{figure}[t]
    \centering
    \includegraphics[width=8.2cm, trim=10 10 10 10,clip]{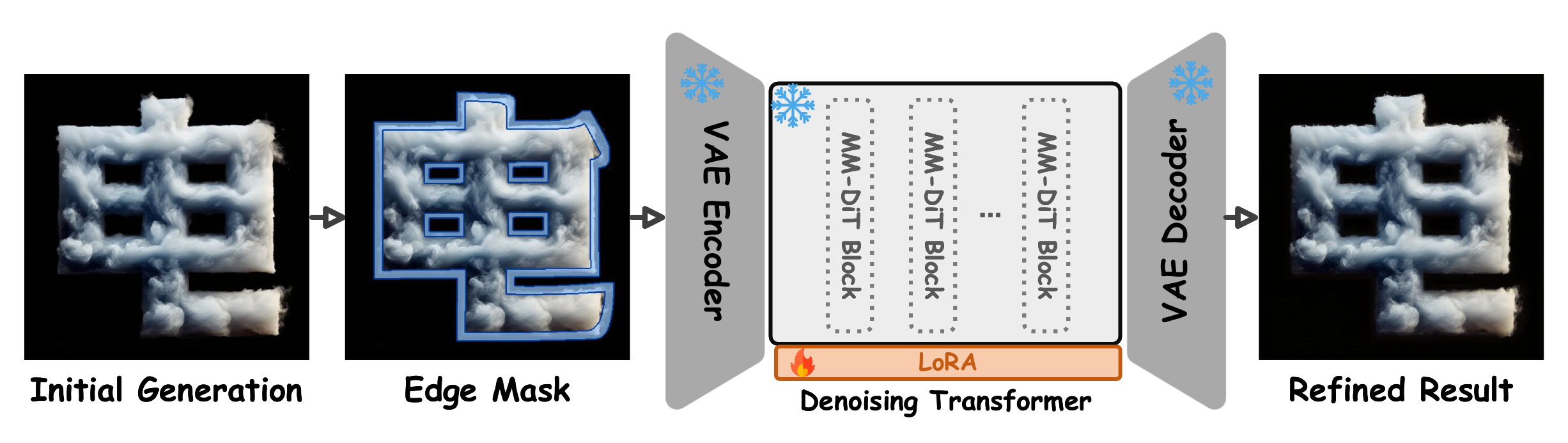}
    \caption{Edge repainting model.}
    \label{fig:stage2}
\end{figure}

\noindent
\textbf{Edge Repainting for Boundary Refinement.}
The proposed model generates visually realistic and stylistically consistent glyphs. However, for some styles, particularly those with amorphous elements, boundaries often appear overly smooth, lacking natural variation. This limitation stems from the fact that glyph masks used during inference are derived from standard font libraries, which possess uniform and clean contours. When applied to elements without fixed structures, such as clouds, the model rigidly adheres to mask boundaries, yielding outputs that diverge from their natural appearance and user expectations. Ideally, glyph boundaries should reflect the intrinsic characteristics of the reference element rather than conforming to overly regular shapes. To mitigate this issue, we introduce an edge repainting module as an optional post-processing step to restore lost stylistic boundary details. Leveraging the observation that reference samples typically exhibit style-specific edge patterns, we fine-tune a pre-trained FLUX.1-Fill model via LoRA to serve as a dedicated boundary refinement network. As illustrated in Figure \ref{fig:stage2}, a narrow mask region is defined along the glyph contours, where the model is tasked with reconstructing these regions guided by the surrounding visual context. This enables the generation of boundaries that align more closely with the reference style, thereby enhancing both the visual quality and stylistic fidelity of the final output.

\section{Experiments}
\label{sec:formatting}

\begin{figure}[t]
    \centering
    \includegraphics[width=8.3cm, trim=10 10 10 10,clip]{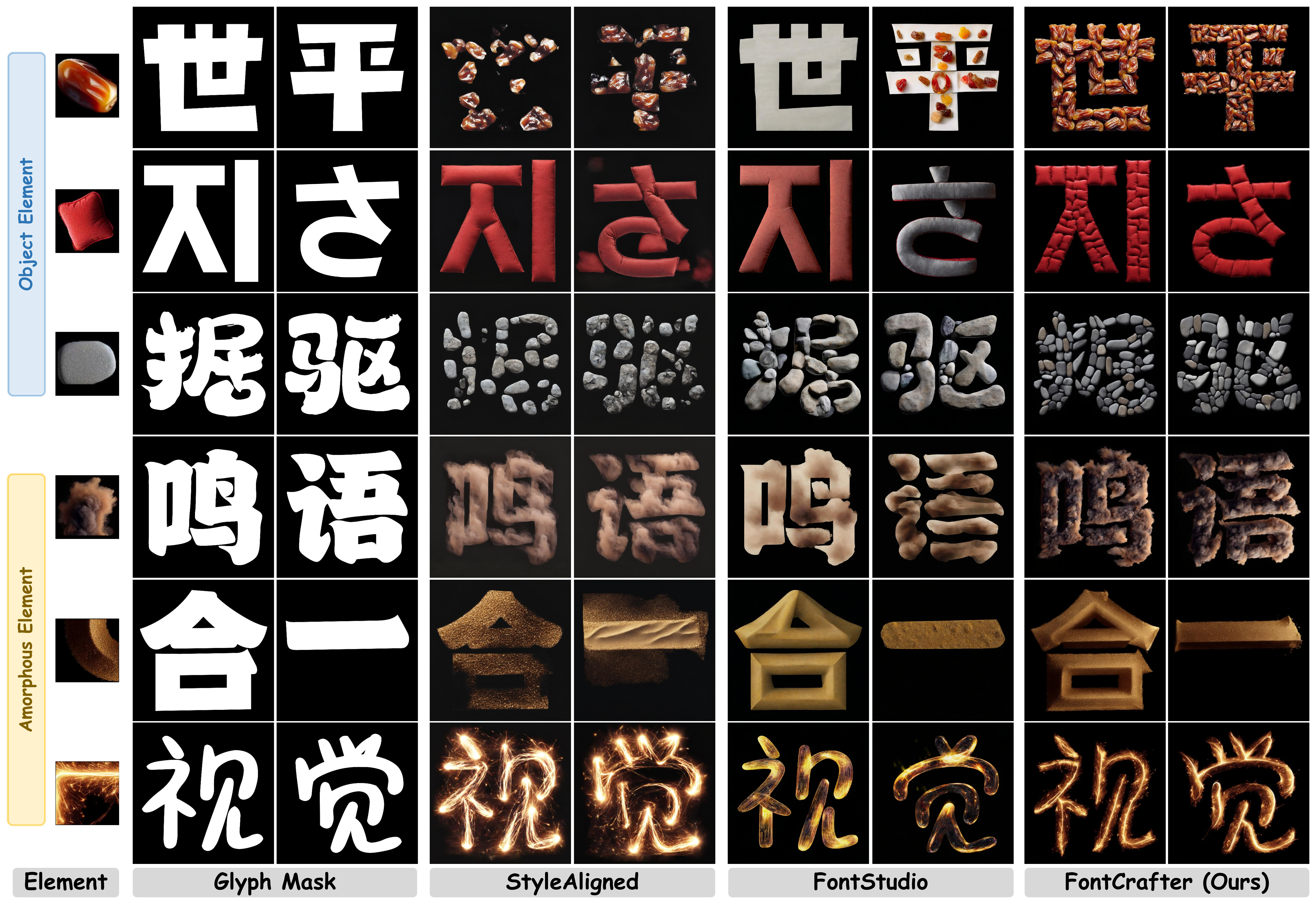}
    \caption{Visual comparison with zero-shot methods.}
    \label{fig:exp_sota}
\end{figure}

\begin{figure}[t]
    \centering
    \includegraphics[width=7.7cm, trim=10 10 10 10,clip]{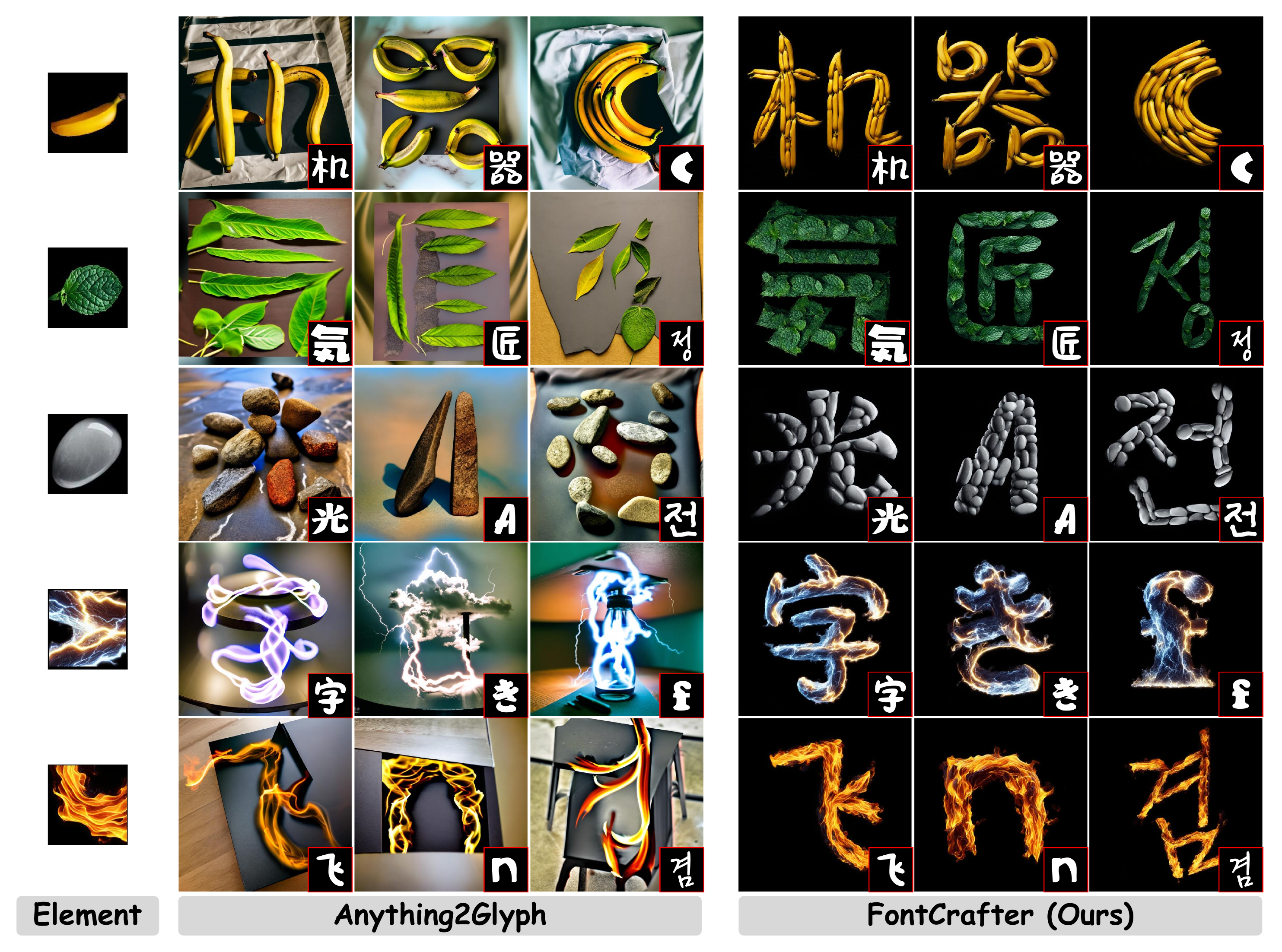}
    \caption{Visual comparison with Anything2Glyph.}
    \label{fig:exp_sota_a2g}
\end{figure}

\begin{table}[t]
\caption{Quantitative comparison with state-of-the-art methods. (O. denotes object element, and A. denotes amorphous element. $\uparrow$: Higher is better; $\downarrow$: Lower is better)}
\renewcommand{\arraystretch}{1.2}
\centering
\footnotesize
\setlength{\tabcolsep}{0.9mm}{
\begin{tabular}{lccccccc}
\Xhline{1pt}
\multicolumn{1}{c}{\multirow{2}{*}{Method}} & \multirow{2}{*}{Type}      & \multirow{2}{*}{FID$\downarrow$} & \multirow{2}{*}{CLIP\textsubscript{Im}$\uparrow$} & \multirow{2}{*}{FID\textsubscript{p}$\downarrow$} & \multicolumn{2}{c}{User Study} & \multirow{2}{*}{SR$\uparrow$} \\ \cline{6-7}
\multicolumn{1}{c}{}                        &                            &                      &                       &                        & Cons.$\uparrow$          & Rd.$\uparrow$          &                      \\ \hline
StyleAligned                                & \multirow{3}{*}{O.}    & 200.3                & 0.70                  & 291.2                  & 73.2           & 78.8          & 2.5                  \\
FontStudio                                  &                            & 205.4                & 0.75                  & 271.3                  & 72.6           & 80.6          & 4.0                  \\
Ours                                        &                            & \textbf{127.5}                & \textbf{0.91}                  & \textbf{190.6}                  & \textbf{92.0}           & \textbf{94.2}          & \textbf{93.5}                 \\ \hline
StyleAligned                                & \multirow{3}{*}{A.} & 227.9                & 0.74                  & 304.2                  & 85.2           & 82.6          & 4.0                  \\
FontStudio                                  &                            & 225.2                & 0.73                  & 283.1                  & 84.8           & 89.4          & 6.5                  \\
Ours                                        &                            & \textbf{128.3}                & \textbf{0.92}                  & \textbf{193.4}                  & \textbf{96.6}           & \textbf{92.4}          & \textbf{89.5}                 \\ \hline\hline
Anything2Glyph                              & \multirow{2}{*}{-}         & 297.8                & 0.33                  & 372.1                  & 42.6           & 45.2          & 1.5                  \\
Ours                                        &                            & \textbf{213.6}                & \textbf{0.91}                  & \textbf{221.5}                  & \textbf{92.8}           & \textbf{93.8}          & \textbf{98.5}                 \\ \bottomrule
\end{tabular}}
\label{tab:sota}
\end{table}

\subsection{Comparison with Zero-Shot Methods}
\textbf{Baselines.}
We evaluate our method against two diffusion-based baselines capable of zero-shot generation:
(1) FontStudio, a recent state-of-the-art method designed for artistic font generation; and
(2) StyleAligned, a versatile style transfer approach integrated with ControlNet to ensure structural guidance. 
For a fair comparison, both baselines are re-trained on our dataset using their publicly available implementations.
The test set comprises 100 unique styles from both amorphous and object elements, all unseen during training.

\noindent
\textbf{Qualitative Results.}
Figure \ref{fig:exp_sota} presents visual comparisons across multiple languages. Our method excels in several key aspects: 
(1) Texture fidelity: Baselines often fail to capture fine-grained style details, resulting in noticeable differences in surface texture and color. In contrast, our approach accurately transfers these style features. 
(2) Structural fidelity: Existing methods struggle to preserve the structure of reference elements, particularly for object elements, which are often degraded to simple patterns. Our method faithfully maintains the shape and internal structure of each object instance. 
(3) Glyph structure alignment: Baselines may produce unreadable glyphs for complex logographic characters, whereas our model generates glyphs well-aligned with the input masks.

\noindent
\textbf{Quantitative Evaluation.}
We evaluate all methods using a comprehensive set of metrics: 
(1) Given the original image $I_{\text{ori}}$ from the test set, we perform glyph augmentation to obtain $I_{\text{aug}}$, and generate $I_{\text{gen}}$. We compute \textit{FID} \cite{heusel2017gans} to assess the distributional distance between generated results and ground truth, and CLIP image similarity (\textit{CLIP\textsubscript{Im}}) \cite{radford2021learning} to measure per-sample differences. 
(2) To evaluate fidelity to the reference elements, we employ patch-level FID metric (\textit{FID\textsubscript{p}}). Specifically, for a test image $I_{\text{ori}}$, we generate a cross-lingual glyph $I_{\text{cross}}$ using the same elements. We then randomly sample image patches around the glyph edges of $I_{\text{ori}}$ or $I_{\text{cross}}$, and compute their FID scores. These patches primarily capture element-level features while ignoring glyph shape, thereby reflecting style consistency. 
(3) We conduct a user study to evaluate consistency and readability. Consistency \textit{(Cons.)} requires participants to judge whether the generated fonts are consistent with the reference elements in both texture and structure. Readability \textit{(Rd.)} measures whether the generated glyphs can be correctly recognized without stroke errors. In total, we collect 500 responses from participants. 
(4) We sample 200 multilingual generated images from different methods and prompt GPT to select preferred outputs, measuring the selection rate (\textit{SR}) of each method.
The quantitative results in Table \ref{tab:sota} show that our method consistently outperforms others across multiple metrics.

\subsection{Comparison with Anything2Glyph}
We compare our method with Anything2Glyph, which employs text prompts as style control and is capable of generating glyphs composed of objects. We evaluate it on the same style categories defined in their pre-set collection. As shown in Figure \ref{fig:exp_sota_a2g}, Anything2Glyph often produces cluttered backgrounds and struggles to adapt to complex glyph structures, resulting in unrecognizable outputs. In contrast, our method generates clean backgrounds while accurately preserving glyph shapes. 
Moreover, relying solely on text limits Anything2Glyph to coarse control over object categories, leading to inconsistent styles. Our method, instead, leverages reference elements to provide fine-grained control and ensure stylistic consistency across glyphs. The quantitative results in Table \ref{tab:sota} further confirm the superiority of our method.

\begin{figure}[t]
    \centering
    \includegraphics[width=8.2cm, trim=10 10 10 10,clip]{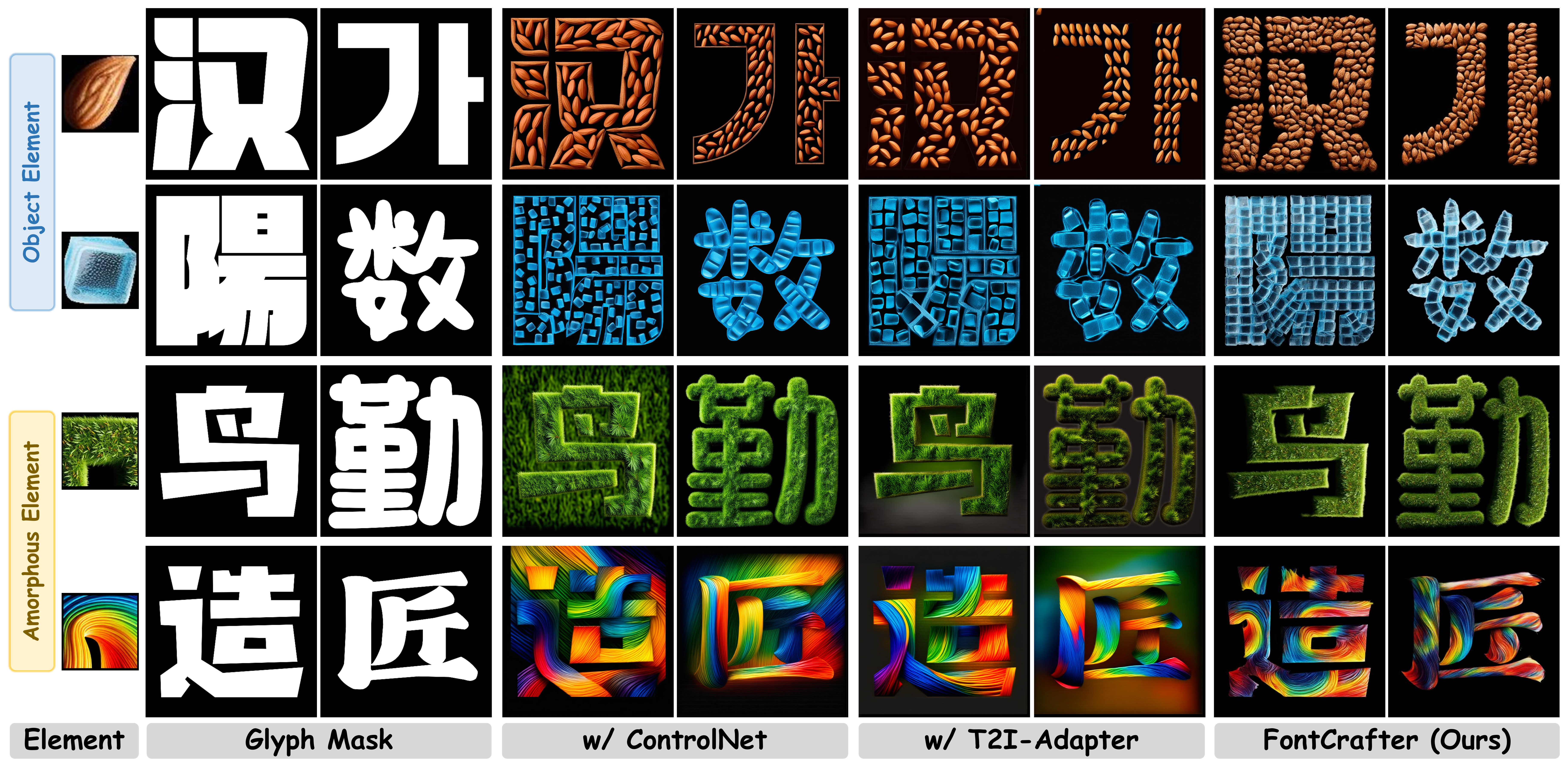}
    \caption{Visual comparison of different glyph control methods.}
    \label{fig:exp_controlnet_t2iadapter}
\end{figure}

\begin{figure}[t]
    \centering
    \includegraphics[width=5.3cm, trim=10 10 10 10,clip]{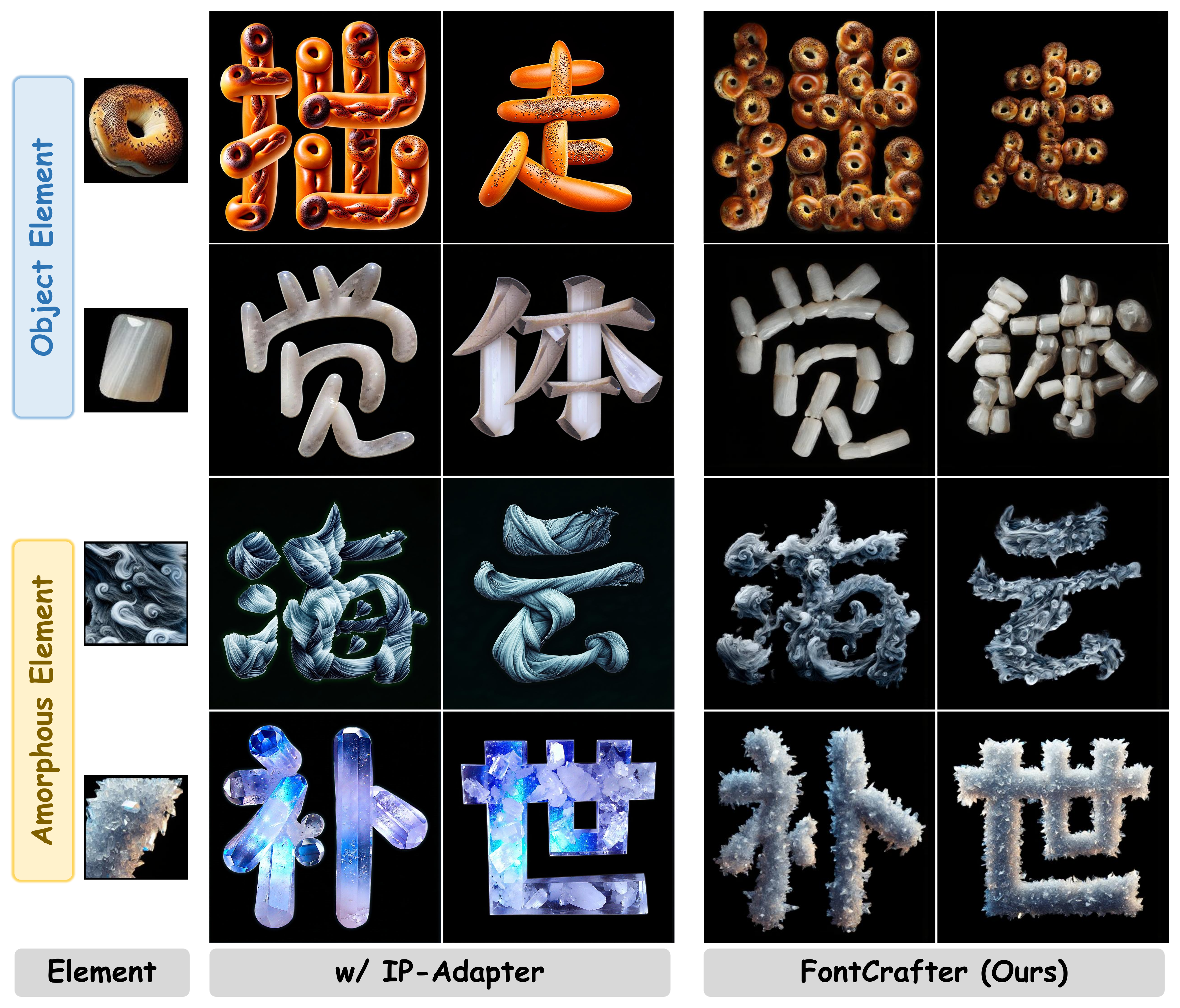}
    \caption{Comparison of in-context control with IP-Adapter.}
    \label{fig:exp_ipa}
\end{figure}

\begin{table}[]
\caption{Quantitative results of the ablation study.}
\renewcommand{\arraystretch}{1.2}
\centering
\footnotesize
\setlength{\tabcolsep}{0.7mm}{
\begin{tabular}{lccccccc}
\Xhline{1pt}
\multicolumn{1}{c}{\multirow{2}{*}{Method}} & \multirow{2}{*}{Type} & \multirow{2}{*}{\#Params} & \multirow{2}{*}{FID$\downarrow$} & \multirow{2}{*}{CLIP\textsubscript{Im}$\uparrow$} & \multirow{2}{*}{FID\textsubscript{p}$\downarrow$} & \multicolumn{2}{c}{User Study} \\ \cline{7-8} 
\multicolumn{1}{c}{}                        &                       &                           &                      &                       &                        & Cons.$\uparrow$          & Rd.$\uparrow$          \\ \hline
w/ ControlNet                               & \multirow{5}{*}{O.}   & 743.81M                  & 193.2                & 0.74                  & 252.1                  & 68.4           & 82.2          \\
w/ T2I-Adapter                              &                       & 79.03M                   & 183.1                & 0.75                  & 246.2                  & 81.2           & 86.8          \\
w/ IP-Adapter                               &                       & -                         & 213.2                & 0.71                  & 283.2                  & 62.2           & 89.0          \\
Ours                                        &                       & 22.4M                      & \textbf{127.5}                & \textbf{0.91}                  & \textbf{190.6}                  & \textbf{92.0}           & \textbf{94.2}          \\ \hline
w/ ControlNet                               & \multirow{5}{*}{A.}   & 743.81M                  & 197.1                & 0.77                  & 247.9                  & 71.2           & 83.4          \\
w/ T2I-Adapter                              &                       & 79.03M                   & 189.5                & 0.79                  & 231.4                  & 84.0           & 86.2          \\
w/ IP-Adapter                               &                       & -                         & 193.1                & 0.75                  & 293.1                  & 68.4           & 86.0          \\
Ours                                        &                       & 22.4M                      & \textbf{128.3}                & \textbf{0.92}                  & \textbf{193.4}                  & \textbf{96.6}           & \textbf{92.4}          \\ \bottomrule
\end{tabular}}
\label{tab:ablation}
\end{table}

\subsection{Ablation Study}

\noindent
\textbf{Ablation on glyph control.} We replace the proposed CMA with two classical spatial control methods: ControlNet \cite{zhang2023adding} and T2I-Adapter \cite{mou2024t2i}.  As shown in Figure \ref{fig:exp_controlnet_t2iadapter}, this substitution degrades performance in several aspects:
(1) over-constrained mask geometry with undesirable edge artifacts; 
(2) reduced consistency between generated styles and reference elements;
(3) style texture leakage into background regions;
(4) substantially higher trainable parameter counts compared to our lightweight CMA.
The quantitative results are provided in Table \ref{tab:ablation}.

\noindent
\textbf{Ablation on style control.}
We employ IP-Adapter \cite{ye2023ip} for style control instead of our visual in-context generation. As shown in Figure \ref{fig:exp_ipa}, IP-Adapter offers only coarse-grained control: generated glyphs capture color and category-level traits but fail to preserve fine-grained textures and structural details. In contrast, our method accurately transfers fine-grained element features, producing glyphs fully composed of the reference elements. Quantitative results in Table \ref{tab:ablation} further show that our method significantly outperforms IP-Adapter, especially in style consistency.

\begin{figure}[t]
    \centering
    \includegraphics[width=7.4cm, trim=10 10 10 10,clip]{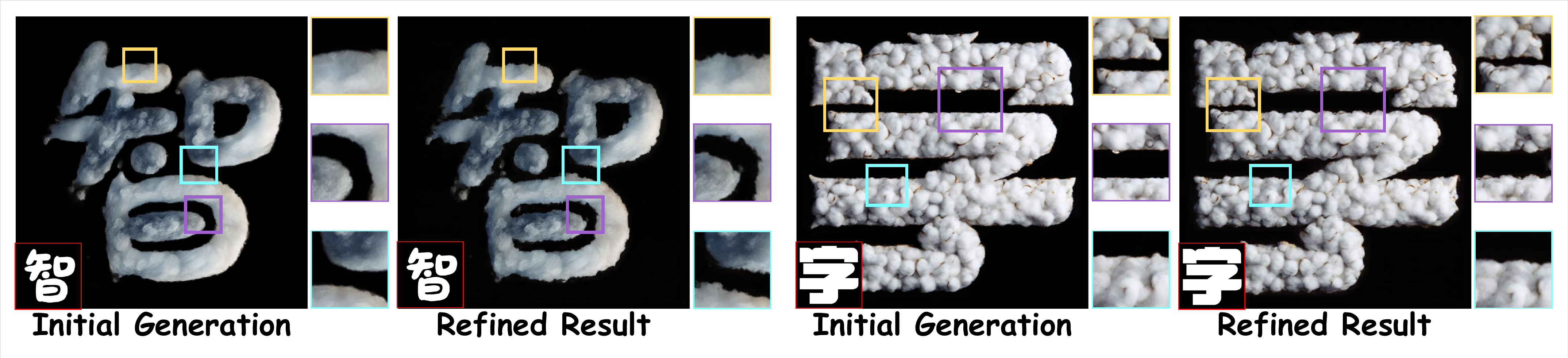}
    \caption{Edge repainting effect on glyph contours.}
    \label{fig:exp_edge_refinement}
\end{figure}

\noindent
\textbf{Effect of Edge Repainting.}
To address overly smooth contours observed in some initial generations, we introduce an edge repainting model as an optional post-processing step. This module refines glyph boundaries by reconstructing edge regions. As shown in Figure \ref{fig:exp_edge_refinement}, the refined results remove unnatural, rigid outlines and produce more natural, visually rich edge patterns that better reflect the reference element’s style, such as the wispy textures of clouds. This results in improved visual realism and stylistic fidelity.

\begin{figure}[t]
    \centering
    \includegraphics[width=7.0cm, trim=10 40 10 10,clip]{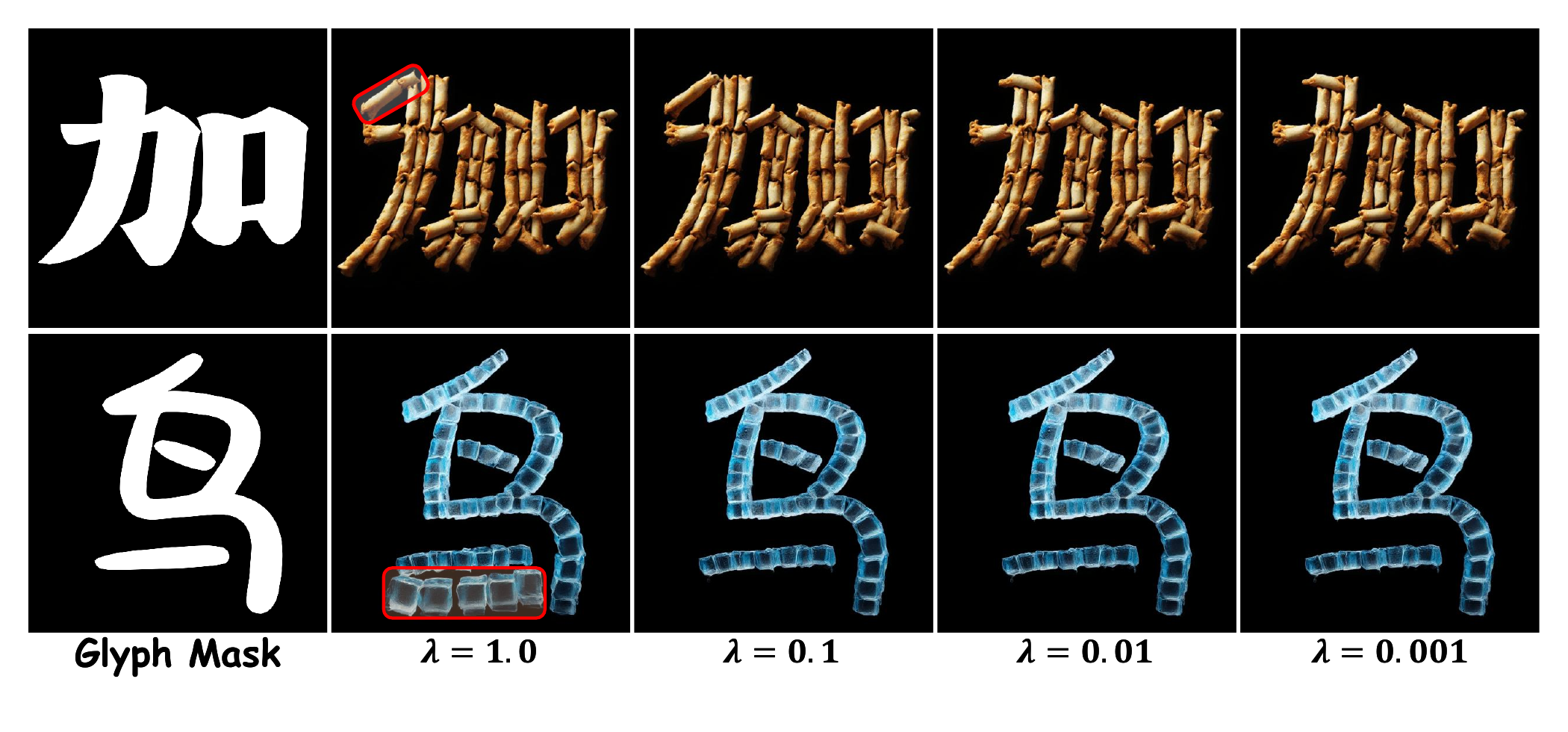}
    \caption{Visual results of dehallucination.}
    \label{fig:exp_stroke_correction}
\end{figure}

\noindent
\textbf{Effectiveness of dehallucination.}
Figure \ref{fig:exp_stroke_correction} illustrates the effect of attention redirection on dehallucination.  When the suppression factor $\lambda$ is set to 1, the mechanism is disabled, and the generated results contain spurious strokes. As $\lambda$ decreases, these unwanted strokes are progressively removed while the intended strokes remain intact. This demonstrates the effectiveness of attention redirection in selectively attenuating attention between potentially interfering regions without affecting the rest of the image.

\begin{figure}[t]
    \centering
    \includegraphics[width=7.0cm, trim=10 10 10 10,clip]{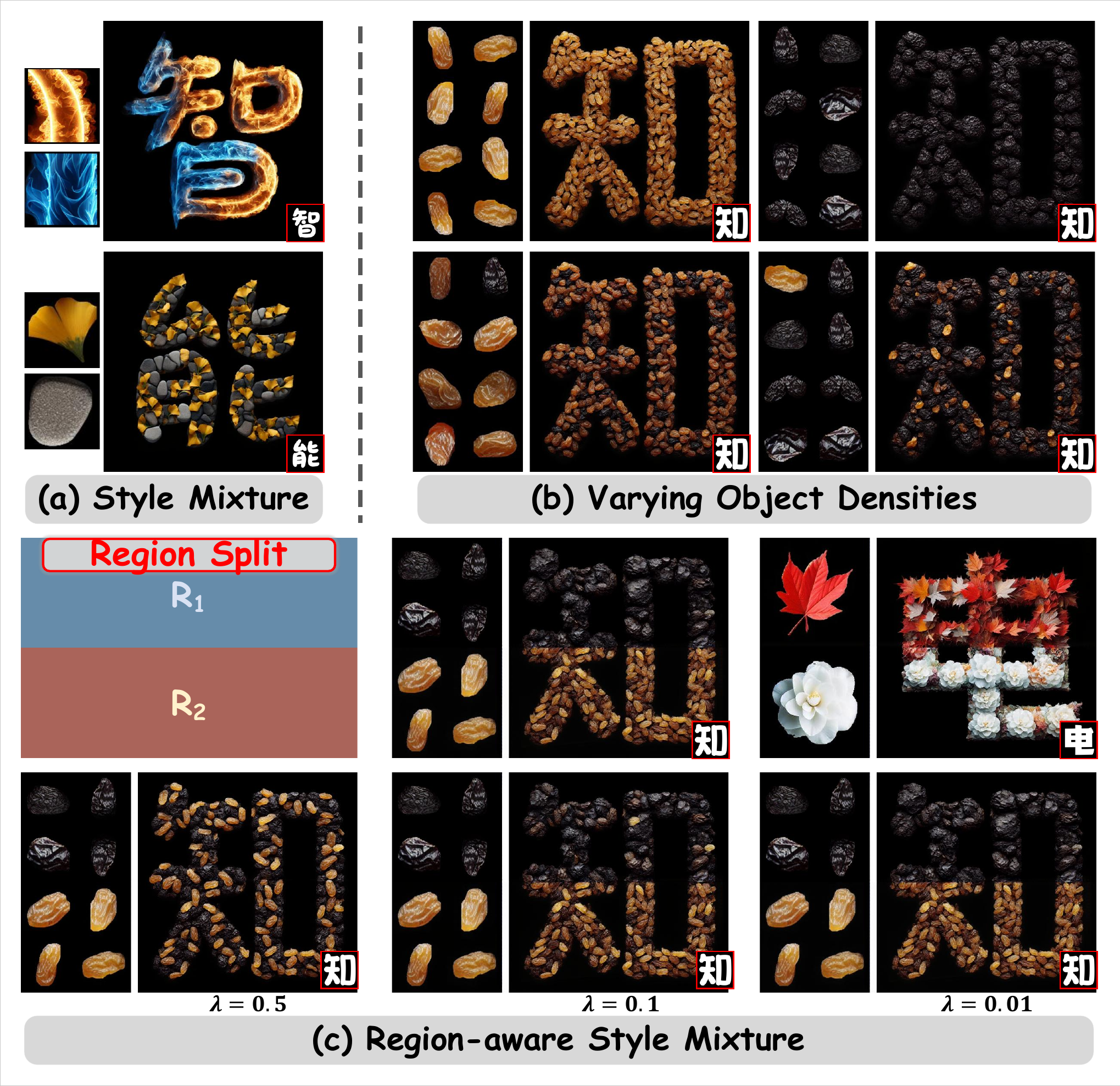}
    \caption{Visual results of style mixture.}
    \label{fig:exp_style}
\end{figure}

\subsection{Style Mixture}
Our visual in-context generation strategy controls output style via a constructed reference region. It naturally supports cross-category style mixing by simply pasting multiple reference images. As shown in Figure \ref{fig:exp_style}(a), when two distinct textures or objects are combined, the generated glyphs integrate features from both styles, yielding visually coherent results. Additional examples can be found in Figure \ref{fig:teaser}. Furthermore, our framework allows fine-grained control over style composition by adjusting the density of each element within the reference region. As illustrated in Figure \ref{fig:exp_style}(b), the proportions of orange and black raisins in the generated glyphs closely match their densities in the reference region. This demonstrates the model’s ability to translate local style distributions into the output, providing users with flexible and intuitive control over the visual appearance.

We further employ the attention redirection mechanism to enable region-aware style mixing. As illustrated in Figure \ref{fig:exp_style}(c), the input is divided into two horizontal regions, $R_1$ and $R_2$, spanning both the reference and glyph regions. Attention suppression is then applied to reduce cross-region interactions. Consequently, the upper part of the glyph primarily attends to objects in $R_1$, while the lower part attends to $R_2$, thereby achieving region-aware style control.
We also investigate the effect of the suppression factor $\lambda$. As shown in the second row of Figure \ref{fig:exp_style}(c), reducing $\lambda$ strengthens regional isolation, resulting in more distinct and well-separated style distributions across the glyph.

\begin{figure}[t]
    \centering
    \includegraphics[width=8.5cm, trim=25 10 10 10,clip]{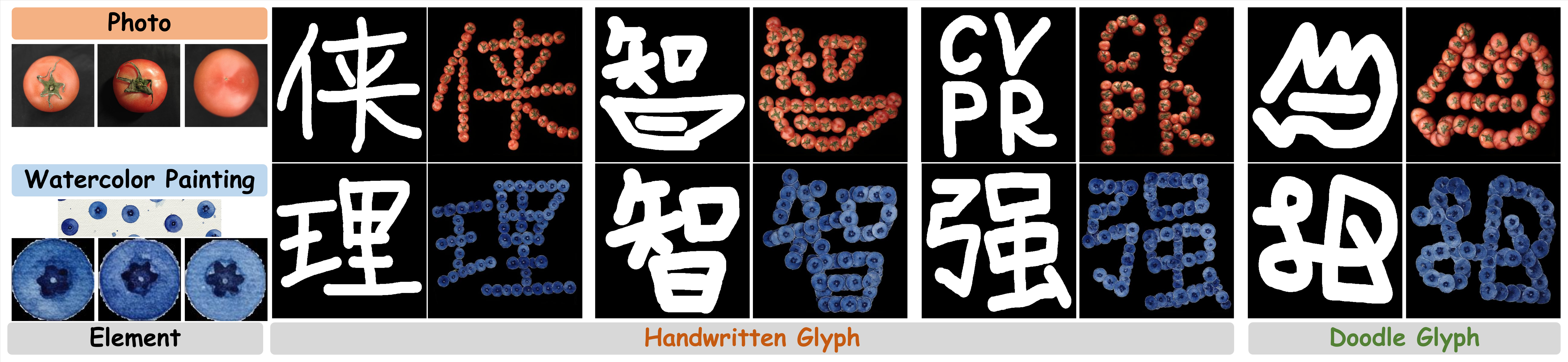}
    \caption{Generalization to real-world elements and glyphs.}
    \label{fig:exp_user}
\end{figure}

\subsection{Generalization to Real-World Scenarios}

To assess the robustness of our framework, we evaluate it on diverse real-world scenarios beyond synthetic data. As shown in Figure~\ref{fig:exp_user}, our method effectively handles varied element references, including self-captured photographs and artistic watercolors, and demonstrates strong structural adaptability on complex, non-standard glyphs such as hand-drawn doodles and handwritten characters.  
This generalization arises from two key factors. First, the diversity of the ElementFont dataset exposes the model to rich variations in element appearance and structure, enhancing robustness to unseen inputs. Second, the visual in-context generation strategy directly conditions on reference elements, facilitating the transfer of fine-grained visual characteristics.
\section{Conclusion}
\label{sec:rationale}

In this paper, we propose FontCrafter, an element-driven framework for controllable artistic font generation via visual in-context synthesis. Our method can produce glyphs that faithfully preserve texture and structural details of the reference elements, while enabling flexible and controllable style mixing. To support this task, we construct ElementFont, a large-scale dataset featuring diverse styles composed of amorphous and object elements. ElementFont offers a comprehensive benchmark for artistic font creation and facilitates future research in this area. 

\noindent
\textbf{Acknowledgement} This work was supported by National Natural Science Foundation of China (Grant No. 62506061), Shanghai Key Laboratory of Intelligent Information Processing, Fudan University (Grant No. IIPL-2025-RD4-01), and Fundamental Research Funds for the Central Universities (Grant No. DUT25YG207). We thank Zimei Li, Hua Zhong, Jiayan He, and Tengbo Pan for their contributions to the construction of the dataset.

{
    \small
    \bibliographystyle{ieeenat_fullname}
    \bibliography{main}
}


\end{document}